\documentclass[letterpaper, 10 pt, conference]{ieeeconf}  

\IEEEoverridecommandlockouts                              

\usepackage{amsfonts,amssymb,amsmath,bm}
\usepackage{amstext}
\usepackage{glossaries}
\usepackage{graphicx}
\usepackage[noend]{algpseudocode}
\usepackage[symbol]{footmisc}
\usepackage{booktabs}
\usepackage[font=footnotesize,labelfont=bf,tableposition=top]{caption}
\usepackage{multicol}
\usepackage{blindtext}
\usepackage[normalem]{ulem}
\usepackage{xcolor}
\usepackage[hidelinks]{hyperref}
\usepackage{cite}
\usepackage{subcaption}
\usepackage[capitalise]{cleveref}

\newacronym{rl}{RL}{Reinforcement Learning}
\newacronym{drl}{DRL}{Deep Reinforcement Learning}
\newacronym{hrl}{HRL}{Hierarchical Reinforcement Learning}
\newacronym{saferl}{SafeRL}{Safe Reinforcement Learning}
\newacronym{avi}{AVI}{Approximate Value-Iteration}
\newacronym{api}{API}{Approximate Policy-Iteration}
\newacronym[plural=MDPs, firstplural=Markov Decision Processes (MDPs)]{mdp}{MDP}{Markov Decision Process}
\newacronym{cmdp}{CMDP}{Constrained Markov Decision Processes}
\newacronym{safeexp}{SafeExp}{Safe Exploration}
\newacronym{kl}{KL}{Kullback-Leibler Divergence}
\newacronym{gae}{GAE}{Generalized Advantage Estimation}
\newacronym{papi}{PAPI}{Projections for Approximate Policy Iteration}
\newacronym{her}{HER}{Hindsight Experience Replay}
\newacronym{ham}{HAM}{Hierarchy of Abstract Machines}
\newacronym{mom}{MOM}{Measure of Manipulability}
\newacronym{bo}{BO}{Bayesian Optimization}
\newacronym{hebo}{HEBO}{Heteroscedastic Evolutionary Bayesian Optimisation}
\newacronym{ucb}{UCB}{Upper Confidence Bound}
\newacronym{pi}{PI}{Probability of Improvement}
\newacronym{ei}{EI}{Expected Improvement}
\newacronym{nl}{NLP}{Nonlinear Programming}
\newacronym{lp}{LP}{Linear Programming}
\newacronym{qp}{QP}{Quadratic Programming}
\newacronym{aqp}{AQP}{Anchored Quadratic Programming}
\newacronym{ode}{ODE}{Ordinary Differential Equation}
\newacronym{atacom}{ATACOM}{\underline{A}cting on the \underline{TA}ngent Space of the \underline{CO}nstraint \underline{M}anifold}
\newacronym{ivp}{IVP}{Initial Value Problem}
\newacronym{rref}{RREF}{Reduced Row Echlon Form}
\newacronym{rcef}{RCEF}{Reduced Column Echlon Form}
\newacronym{cpo}{CPO}{Constrained Policy Optimization}
\newacronym{trpo}{TRPO}{Trust Region Policy Optimization}
\newacronym{rmp}{RMP}{Riemannian Motion Policies}
\newacronym{dnn}{DNN}{Deep Neural Networks}
\newacronym{sdf}{SDF}{Signed Distance Function}
\newacronym{redsdf}{ReDSDF}{Regularized Deep Signed Distance Fields}
\newacronym{apf}{APF}{Artificial Potential Fields}
\newacronym{hri}{HRI}{Human-Robot Interaction}
\newacronym{salem}{SaLeM}{Safe Exploration on Learned Manifolds}
\newacronym{poi}{PoI}{Point of Interest}

\def\EV{\mathbb{E}}
\def\RR{\mathbb{R}}

\def\vzero{{\bm{0}}}

\def\vmu{{\bm{\mu}}}

\def\valpha{{\bm{\alpha}}}

\def\va{{\bm{a}}}

\def\vc{{\bm{c}}}

\def\vf{{\bm{f}}}

\def\vp{{\bm{p}}}
\def\vq{{\bm{q}}}

\def\vs{{\bm{s}}}

\def\vu{{\bm{u}}}

\def\vx{{\bm{x}}}

\def\mF{{\bm{F}}}
\def\mG{{\bm{G}}}

\def\mI{{\bm{I}}}
\def\mJ{{\bm{J}}}

\def\mN{{\bm{N}}}

\def\mZ{{\bm{Z}}}

\newcommand\scalemath[2]{\scalebox{#1}{\mbox{\ensuremath{\displaystyle #2}}}}

\title{\LARGE \bf
Safe Reinforcement Learning of Dynamic High-Dimensional\\ Robotic Tasks: Navigation, Manipulation, Interaction}

\author{Puze Liu$^{1}$, Kuo Zhang$^{1}$, Davide Tateo$^{1}$, Snehal Jauhri$^{1}$, Zhiyuan Hu$^{1}$, Jan Peters$^{1-4}$ and Georgia Chalvatzaki$^{1,3}$
\thanks{This work is supported by the China Scholarship Council (No. 201908080039), the DFG Emmy Noether Programme (CH 2676/1-1), the Daimler-Benz Foundation, and the Hessian Competence Center for High Performance Computing – funded by the Hessen State Ministry of Higher Education, Research and the Arts, and partially supported by the German Federal Ministry of Education and Research (BMBF) within 
the collaborative KIARA project (grant no. 13N16274). \newline
$^{1}$ Computer Science Department, Technical University Darmstadt, $^{2}$~German Research Center for AI (DFKI), Research Department: Systems AI for Robot Learning, $^{3}$~Hessian.AI, $^{4}$~Centre for Cognitive Science.
{\tt\footnotesize puze@robot-learning.de,\{davide.tateo, snehal.jauhri, jan.peters, georgia.chalvatzaki\} @tu-darmstadt.de}}
}

\begin{document}
\let\oldtwocolumn\twocolumn
\renewcommand\twocolumn[1][]{%
    \oldtwocolumn[{#1}{
    \begin{center}
    \vspace{-1.3cm}
    \captionsetup{type=figure}
    \includegraphics[width=0.99\textwidth]{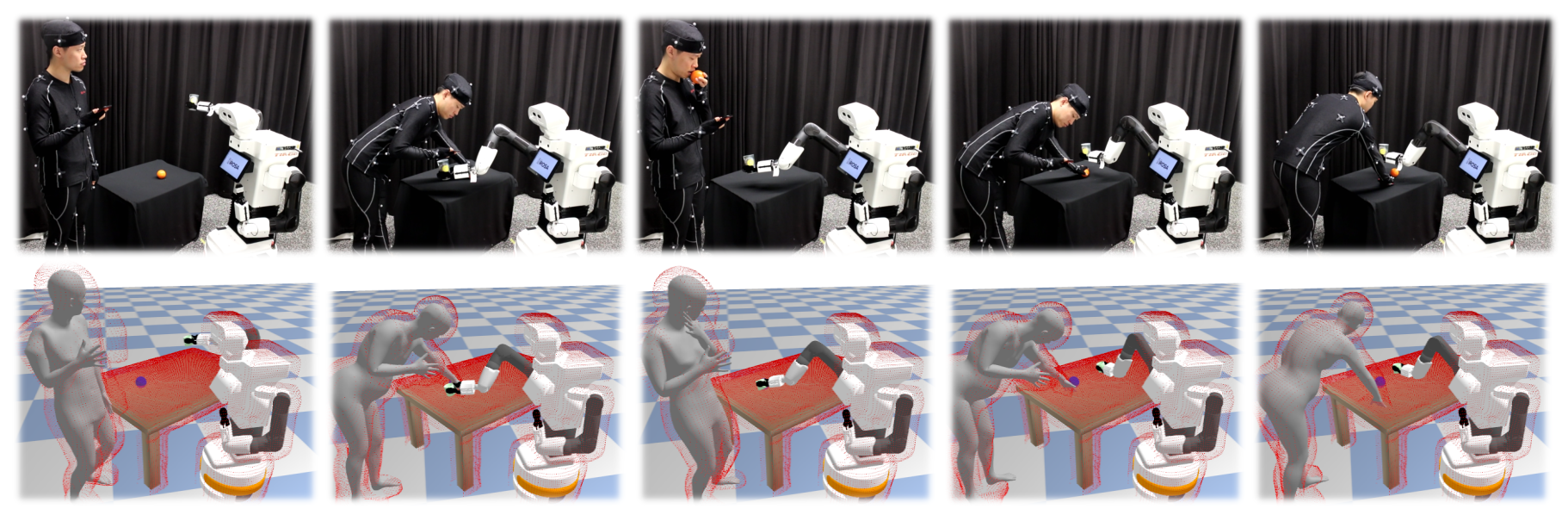}
    \vspace{-0.35cm}
    \captionof{figure}{\textit{Top:} Sequence of real-world human-robot interaction. The human casually looks over the phone ignoring the robot that tries to deliver safely a cup of water. While the human reaches for an orange, the robot smoothly avoids collisions, maintaining the glass in an upward position. \textit{Bottom:} The simulated digital twin of the real scene that illustrates the signed distance fields of 0.1m (red points) of the human, the robot, and the table, that are used as constraint models in our Safe Reinforcement Learning algorithm.}\vspace{-0.15cm}
    \label{fig:main_figure}
        \end{center}
    }]
}

\maketitle
\thispagestyle{empty}
\pagestyle{empty}

\begin{abstract}
Safety is a fundamental property for the real-world deployment of robotic platforms. Any control policy should avoid dangerous actions that could harm the environment, humans, or the robot itself. In reinforcement learning (RL), safety is crucial when exploring a new environment to learn a new skill. This paper introduces a new formulation of safe exploration for Robot Reinforcement Learning in the tangent space of the constraint manifold that effectively transforms the action space of the RL agent for always respecting safety constraints locally. We show how to apply this approach to a wide range of robotic platforms and how to define safety constraints that represent dynamic articulated objects like humans in the context of robotic RL. Our proposed approach achieves state-of-the-art performance in simulated high-dimensional and dynamic tasks while avoiding collisions with the environment. We show safe real-world deployment of our learned controller on a TIAGo++ robot, achieving remarkable performance in manipulation and human-robot interaction tasks.
\end{abstract}

\section{Introduction}
Safe deployment of general-purpose robotic systems in the real world is an overarching goal of safe learning methods~\cite{brunke2022safe}. We envision robots learning complex high-dimensional tasks in dynamic, unstructured environments following the paradigm of Deep \gls{rl}~\cite{ibarz2021train}. In the standard \gls{rl} setting, the agent interacts with an environment of unknown dynamics, collecting experience for learning a suitable behavior (policy). The exploration of a \gls{mdp} for collecting experience can lead to hazardous situations that can damage the robot or the environment~\cite{pecka2015safe}. In this work, we investigate the problem of \textit{safe exploration} in Deep \gls{rl} for different robotic tasks with high-dimensional state and action spaces, such as robotic manipulation. We also focus on environments that can dynamically change, e.g., due to the dynamic behavior of a human in a \gls{hri} setting. 

Safe learning for control and \gls{rl} is a field of increasing interest in light of the different application areas where autonomous systems should operate~\cite{altman1999constrained,mihai2012safe,ammar2015safe,berkenkamp2017safe}. Three main approaches for Safe\gls{rl} are particularly relevant to our work. First is the Safe set approach, whose objective is to keep the agent in the set of states considered safe~\cite{hans2008safe,Garcia2012safe,akametalu2014reachability,pecka2015safe,alshiekh2018safe,wachi2018safe,koller2018learning}. 
The Safe set approaches rely either on a predefined safety set and backup policies or, starting from an initial safe policy, expand the safe set during learning. The second line of work formalizes the problem as a \gls{cmdp}~\cite{altman1999constrained}. The objective of \gls{cmdp} is to learn an optimal policy where the expected (discounted) cumulative constraint violation does not exceed a threshold. To solve the problem under this formulation, many different approaches have been developed based on Lagrangian optimization~\cite{altman1998constrained,achiam2017constrained,stooke2020responsive,ding2021provably,cowen2022samba}, or using an additional reward signal to penalize constraint violations~\cite{liu2020ipo,tessler2019reward}.
More recent approaches~\cite{sootla2022enhancing,sootla2022saute} use state augmentation, which transforms the \gls{cmdp} problem into an unconstrained one by incorporating the constraint information into the policy.

Safe exploration methods address the safety problem at each step when interacting with the environment \cite{berkenkamp2019safe}. In this setting, we define as safe the behavior of an agent that complies with a set of constraints.
These constraints are usually predefined, approximated, learned, or inferred from an initial safe policy. The algorithm directly samples a safe action or modifies one to enforce constraint satisfaction. These approaches are based on either Lyapunov functions~\cite{berkenkamp2017safe,chow2018lyapunov,chow2019lyapunov}, Hamilton-Jacobi reachability~\cite{fisac2018general}, Control Barrier Functions~\cite{cheng2019end,taylor2020learning}, constraint linearization~\cite{dalal2018safe,liu2022robot}  or planning~\cite{hewing2020learning,shao2021reachability}. While some of these methods have already proven to be effective in some controlled tasks~\cite{fisac2018general,taylor2020learning}, the objective of this work is to lay the foundations for extending the applicability of safe exploration techniques to the complexity of real-world constraints for safe robotic \gls{rl}.

One of the most promising ways to impose safe learning is by \gls{atacom}~\cite{liu2022robot}, a method that effectively maps the action space of an \gls{rl} agent to the tangent space of the constraints, i.e., ensuring local constraint satisfaction. To allow such approach to be effectively used for safe exploration in \gls{rl} of high-dimensional robotic tasks of different complexity, like navigation, manipulation, \gls{hri}, we extend \gls{atacom} towards three key directions. \textbf{(a)} We generalize \gls{atacom} to nonlinear affine control systems, describing a wide range of robotic systems. \textbf{(b)} We show how to define complex constraints that involve the uncontrollable environmental state in \gls{atacom} to allow the agent to explore safely even with dynamic objects in the domain. \textbf{(c)} We provide an in-depth discussion about the tangent space of the constraint manifold and show how to increase numerical stability. 

Thanks to the generalized formulation of \gls{atacom}, we can transform the action space of \gls{rl} agents for acting on the constraint manifold of both hand-designed (e.g., joint limits) and learned constraints (e.g., for avoiding contact with humans), resolving various classes of constraints for different tasks (e.g., differential drive constraint), while ensuring the deployability of the learned policy at test time, since \gls{atacom} is always active. We provide experimental evaluations on different robotic simulated tasks for learning navigation, manipulation, and \gls{hri}, and we show the superior performance of the generalized \gls{atacom} agent in terms of safety and task success-rate against representative baselines from the safe exploration literature. Crucially, we demonstrate the real-world applicability of \gls{atacom} in manipulation and \gls{hri} tasks with complex learned constraint manifolds (e.g., human manifold in \cref{fig:main_figure}), showcasing the ability to preserve local safety during deployment without significant performance losses due to sim-to-real gaps.
\\
\noindent\textit{\underline{Related Work in Safe Learning for Robotics}}
\\
Both safety and adaptability concern all robotic tasks, from manipulation \cite{sukhija2022scalable,martinez2015safe}, \& navigation \cite{weerakoon2022terp,bajcsy2019efficient}, to locomotion~\cite{marco2021robot} \& \gls{hri}~\cite{pang2021towards,pandya2022safe, thumm2022provably, schepp2022sara}.
These two requirements are the main focus of the vast literature on Safe\gls{rl}~\cite{brunke2022safe}. Many different formulations and solutions have been proposed to face this problem, e.g., uncertainty-aware model-based \gls{rl} in robot navigation tasks \cite{kahn2017uncertainty,lutjens2019safe, chalvatzaki2019learn}, offline learning for finding unsafe zones and recovery policy for a surgical robot is proposed in~\cite{thananjeyan2021recovery} and for locomotion in~\cite{yang2022safe}. 
Many works adopt the idea of adding a safety layer on the \gls{rl}-policy, that can adapt an unsafe action to a safe one in conjunction with reachability analysis \cite{shao2021reachability,fisac2019bridging}. An optimization layer was used in \cite{pham2018optlayer} to train a robot-reaching task in the real world. An ensemble of policies is used in \cite{kaushik2022safeapt}, from which the most likely safe policy is transferred to a real robot playing air hockey through episodic interaction. An \gls{rl} framework that filters suboptimal actions in the domain of \gls{hri} is introduced in \cite{el2020towards}, but was not demonstrated on a real-world task. In this paper, we introduce a generalized framework for safe exploration in \gls{rl} of robotic tasks that operates on the tangent space of the constraint manifold~\cite{liu2022robot} satisfying both handcrafted and learned constraints~\cite{liu2022robot}. We showcase real-world performance in challenging tasks, like \gls{hri}, demonstrating that our framework can enable safe robot learning of various tasks across different application areas.

\noindent\textit{\underline{Problem Statement}}
\\
\gls{saferl} applies to problems modeled as a \gls{cmdp} \cite{altman1999constrained} defined by the tuple $<\mathcal{S}, \mathcal{A}, P, \gamma, R, \mathcal{C}>$, where $\mathcal{S}$ is the state space, $\mathcal{A}$ is the action space, $P:\mathcal{S}\times \mathcal{A} \times \mathcal{S} \rightarrow [0, 1]$ is the transition kernel, $\gamma \in (0, 1]$ is the discount factor, $R: \mathcal{S} \times \mathcal{A}\rightarrow \RR$ is the reward function, and $\mathcal{C}:=\{c_i:\mathcal{S}\rightarrow \RR|i\in \mathbb{N}\}$ is a set of constraint functions. We approach the problem of safety in \gls{rl} through the glance of \textit{safe exploration} that prevents constraint violations throughout the learning process. Therefore, we formalize our problem as follows,
\begin{align*}
    \scalemath{0.9}{\max_{\pi}} & \scalemath{0.85}{\quad \EV_{\tau \sim \pi} \left[\sum_{t}^{T}\gamma^t r(\vs_t, \va_t)\right]}\\ \scalemath{0.85}{\mathrm{ s.t.}} & \scalemath{0.85}{\quad  c_i(\vs_t) \leq 0, \quad i \in \{1, 2, ..., N\}, \; t \in \{0, 1, \cdots, T\}}
\end{align*}
where $\tau = \{\vs_0, \va_0, \cdots, \vs_T, \va_T\}$ is the trajectory under policy $\pi$, and $\va_t \sim \pi(\cdot, \vs_t)$ is the action sampled from policy $\pi$. The objective is to maximize the discounted cumulative reward, while satisfying all constraints at each step.

\section{Novel Formulation of ATACOM for Mobile Robots and Manipulators}
In this section, we provide a novel and more general formulation of the \gls{atacom} method. Our proposed formulation allows applying \gls{atacom} for learning a wide variety of mobile robotics and manipulation tasks. Moreover, we handle some critical aspects of the original \gls{atacom}, improving its numerical stability, learning performance, and safe-space structure. Finally, we show how to model complex real-world collision-avoidance constraints using learned \gls{sdf}s, that allow safe learning of complex tasks, as in the domain of \gls{hri}. We first introduce the original design of~\gls{atacom}. Then, we introduce our reformulation to it that allows its generalization to a broader class of problems.
\subsection{Original ATACOM Formulation} 
\gls{atacom}~\cite{liu2022robot} is a method for safe exploration in the tangent space of the constraints' manifold. It converts the constrained \gls{rl} problem to a typical unconstrained one, while handling both equality and inequality constraints. This method allows us to utilize any model-free \gls{rl} algorithm, while maintaining the constraints below a designated tolerance. In \gls{atacom}, the state space $\mathcal{S}$ is separated into two sets, the controllable state space $\mathcal{Q} \subset \RR^n$ and the uncontrollable state space $\mathcal{X} \subset \RR^m$, i.e., $\vs = [\vq^\intercal \; \vx^\intercal]^\intercal\in\RR^{n+m}$. \gls{atacom} prescribes that all $k$ constraints are defined on the controllable variable $\vc(\vq)\leq \vzero$, where $\vc:\RR^n\rightarrow\RR^k$ is differentiable. \gls{atacom} constructs a constraint manifold by introducing the slack variable $\vmu \in \RR^k$ into the constraint 
\begin{equation}
\scalemath{0.85}{
    \mathcal{M}= \left\{(\vq, \vmu): \bar{\vc}(\vq, \vmu)=\vc(\vq) + \frac{1}{2}\vmu^2 = \vzero, \vq \in \mathcal{Q}, \vmu \in \RR^k \right\}
    }
    \label{eq:constraint_manifold}
\end{equation}
The tangent-space bases of the constraint manifold are determined by computing the null space $\mN(\vq, \vmu) \in \RR^{(n+k) \times n}$ of the Jacobian matrix $\mJ(\vq, \vmu)=\left[\frac{\partial }{\partial \vq}\bar{\vc}(\vq, \vmu)^{\intercal}, \frac{\partial }{\partial \vmu}\bar{\vc}(\vq, \vmu)^{\intercal}\right]\in \RR^{k \times(n+k)}$. To simplify the notation, we use $\bar{\vc}$, $\mN$, and $\mJ$, without explicitly writing the dependency on the input. The velocity of the controllable state can be determined by
\begin{equation}
    \scalemath{0.85}{\begin{bmatrix}\dot{\vq}\\  \dot{\vmu}\end{bmatrix} = \mN \valpha - K_c \mJ^{\dagger}\bar{\vc}} \label{eq:atacom_action}
\end{equation}
with the action $\valpha \sim \pi(\cdot|\vq, \vx)$ sampled from the policy. The second term on the right-hand side, with the pseudoinverse of the Jacobian $\mJ^{\dagger}$ and the gain $K_c$, is the error correction term that forces the agent to stay on the manifold, and it is necessary when using time discretization. 

\subsection{Safe Exploration with Generalized ATACOM on Nonlinear Affine Control Systems}
The original formulation of \gls{atacom} assumes a holonomic system, i.e., it assumes that we can set an arbitrary derivative of each generalized coordinate describing the mechanical system. However, many robotics systems of interest are subject to non-holonomic constraints, i.e., non-integrable
constraints, such as the differential drive and the bicycle model, which prevent imposing arbitrary velocities or accelerations on the system's state variables.

To extend the applicability of \gls{atacom} to a broader class of systems, we reformulate it for nonlinear affine control systems~\cite{ames2019control}. In this setting, we assume that the system's velocity of the generalized coordinates can be expressed as
\begin{equation}
    \scalemath{0.9}{\dot{\vq} = \vf(\vq) + \mG(\vq)\va},
    \label{eq:affine_control_system}
\end{equation}
with the control action vector $\va$, and two arbitrary (nonlinear) vector functions of the current state variable  $\vf(\vq), \mG(\vq)$.

Following the original \gls{atacom} derivation, we consider the constraint with uncontrollable state, $\vc(\vq, \vx) \leq \vzero$. The constraint manifold can be defined as $\mathcal{M}=\{(\vq, \vx, \vmu):\bar{\vc}(\vq, \vx, \vmu) = \vzero\}$. Assuming that the velocity of the state variables $\vx$ involved in the constraints are known or estimated, and using the dynamical system in \cref{eq:affine_control_system}, we write the time derivative of the constraint function $\bar{\vc}(\vq,\vx,\vmu)$ as
\begin{align}
    \scalemath{0.85}{\frac{d}{dt}\bar{\vc}(\vq,\vx,\vmu)} & \scalemath{0.85}{=\mJ_{\vq}\dot{\vq} + \mJ_{\vx}\dot{\vx} + \mJ_{\vmu}\dot{\vmu}} \nonumber\\
    & \scalemath{0.85}{=\mJ_{\vq}\vf(\vq) + \mJ_{\vq}\mG(\vq)\va + \mJ_{\vx}\dot{\vx} + \mJ_{\vmu}\dot{\vmu}} \nonumber\\
    & \scalemath{0.85}{=\mF(\vq,\vx,\dot{\vx},\vmu)+ \mJ_{\mG}\va + \mJ_{\vmu}\dot{\vmu} },
    \label{eq:affine_control_constraint_diff}
\end{align}
with $\mJ_{\vq} = \frac{\partial }{\partial \vq}\bar{\vc}(\vq, \vx, \dot{\vx}, \vmu)$, $\mJ_{\vx} = \frac{\partial }{\partial \vx}\bar{\vc}(\vq, \vx, \vmu)$, and $\mJ_{\vmu} = \frac{\partial }{\partial \vmu}\bar{\vc}(\vq, \vx, \vmu)$ the Jacobian matrices w.r.t. the $\vq$, $\vx$, and $\vmu$ variables respectively, $\mF(\vq,\vx,\dot{\vx},\vmu)=\mJ_{\vq}\vf(\vq) + \mJ_{\vx}\dot{\vx}$, and $ \mJ_{\mG}=\mJ_{\vq}\mG(\vq)$. Again, we drop the explicit dependency of the variables $\vq$, $\vmu$ and $\vx$ to simplify the notation.

We can now compute the safe action by imposing zero velocity of constraint violation. Setting the right-hand side of \cref{eq:affine_control_constraint_diff} to 0 and solving for $\va$ and $\dot{\vmu}$ we obtain
\begin{equation}
\scalemath{0.9}{
    \begin{bmatrix}\va \\ \dot{\vmu} \end{bmatrix} = \mN_{[\mG, \vmu]}\valpha -\mJ_{[\mG, \vmu]}^\dagger\mF(\vq,\vx,\dot{\vx},\vmu)},
    \label{eq:atacom_affine}
\end{equation}
where $\mJ_{[\mG, \vmu]}$ is the concatenation of the $\mJ_{\mG}$ and $\mJ_\vmu$ matrices, and $\mN_{[\mG, \vmu]}$ is the null space of $\mJ_{[\mG, \vmu]}$.  As done in \gls{atacom}, it is also possible to add an error correction term $-K_c \mJ^{\dagger}\bar{\vc}$ to the applied control action computed in~\eqref{eq:atacom_affine}: this term makes the system more responsive and able to deal with equality constraints. By the special choice of slack variable function discussed in \cref{subsec:improving_atacom}, we can ensure $\mJ_{[ \mG, \vmu ]}$ is full rank and invertible when $\mu$ is well-defined. We will introduce how the null space matrix is obtained in \cref{subsec:improving_atacom}.

With the extended formulation in \cref{eq:atacom_affine}, we can compute the control action for a wide variety of common robotics kinematics. As an example, we will consider the differential drive kinematics, for which the mobile robot can only move forward/backward along the heading direction and rotate around its center axis. 
The differential drive kinematics can be cast as a nonlinear affine control system as follows
\begin{align}
\scalemath{0.85}{
    \vq = \begin{bmatrix}
            x \\
            y \\
            \theta 
          \end{bmatrix}} 
    & &\scalemath{0.85}{\vf(\vq) = \begin{bmatrix}
                    0 \\
                    0 \\
                    0
                  \end{bmatrix}}
    & &\scalemath{0.85}{\mG(\vq) = \begin{bmatrix}
               \cos{\theta} & 0 \\ 
               \sin{\theta} & 0 \\ 
               0 & 1
            \end{bmatrix}}
    & &\scalemath{0.85}{\va = \begin{bmatrix}
             v \\ 
             \omega
          \end{bmatrix}} 
    \label{eq:constr_diff_drive}
\end{align}
with the Cartesian coordinates $x$ and $y$, the current yaw angle $\theta$, the angular velocity $\omega$, and the speed $v$ along heading direction. 
Given this definitions, it is easy to derive a safe control action using \cref{eq:atacom_affine}.
A similar derivation holds for other kinematics models, e.g. the bicycle kinematics.

Our newly proposed formulation presents a clean and general way to handle systems controlled in velocity/acceleration/jerk or any other arbitrary derivative: this objective can be easily achieved by adding all non-controlled derivatives as state variable $\vq$ and defining appropriately the $\vf(\vq)$ and $\mG(\vq)$ functions. 

\subsection{Robust Tangent Space Bases}
\label{subsec:improving_atacom}
An \gls{atacom} agent explores the tangent space of the constraint manifold. Therefore, obtaining smooth and consistent bases of the tangent space is essential for training the \gls{rl} policy. The original method~\cite{liu2022robot} constructs the constraint manifold by introducing the slack variables in quadratic form and determines the unique tangent space bases by QR decomposition of the Jacobian matrix and \gls{rref}. However, this approach may derive inconsistent bases and suffers numerical stability issues. In this work, we further investigate the numerical problems and introduce a new type of slack variable and a new way of determining the tangent space bases that keep the consistency.
\begin{figure}[b]
    \vspace{-0.5cm}
    \centering
    \includegraphics[width=0.45\linewidth, height=3.2cm, trim=1.8cm 1.8cm 1.8cm 1.8cm, clip]{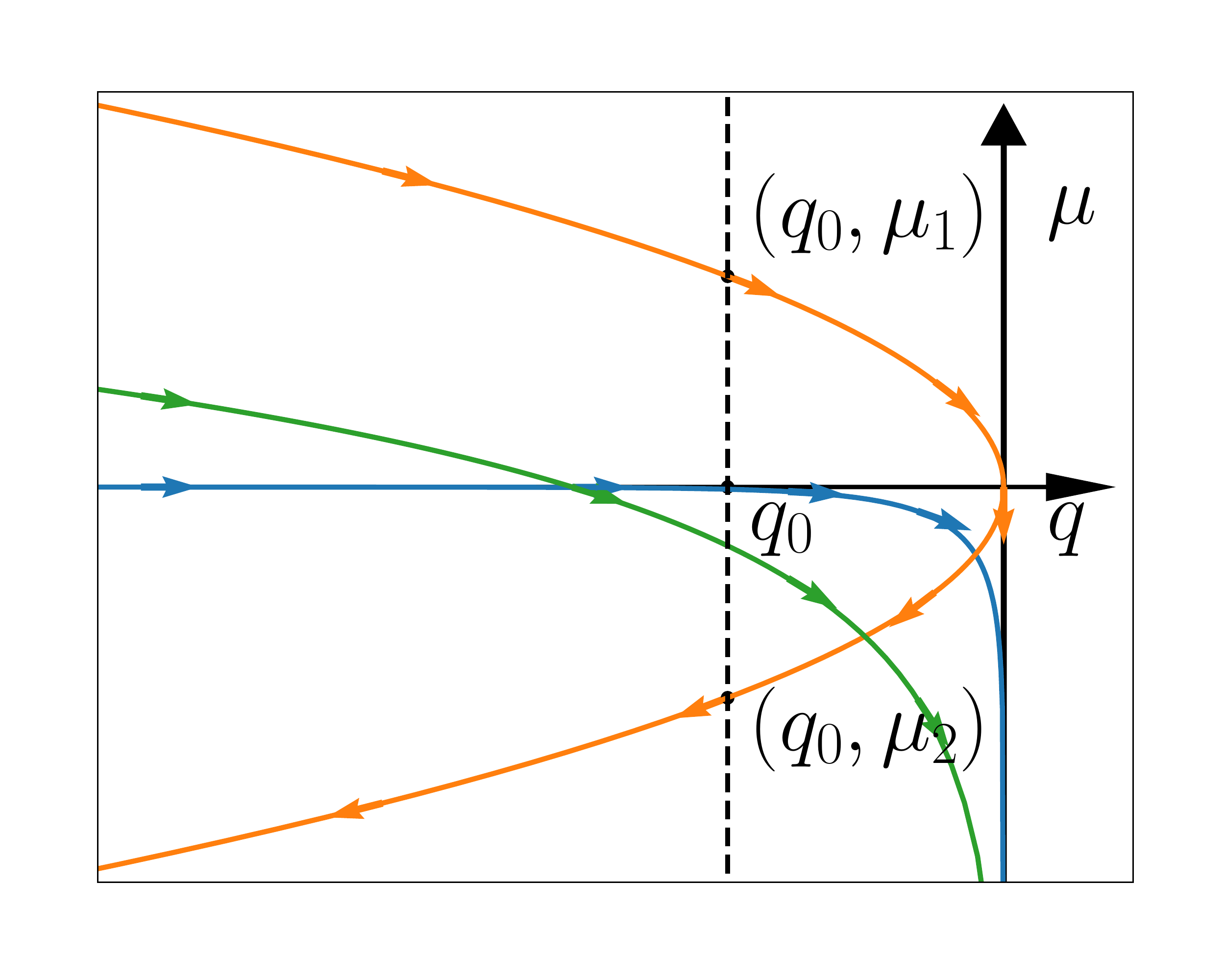}
    \hspace{0.1cm}
    \includegraphics[width=0.45\linewidth, height=3.2cm, trim=1.4cm 1.4cm 1.4cm 1.4cm, clip]{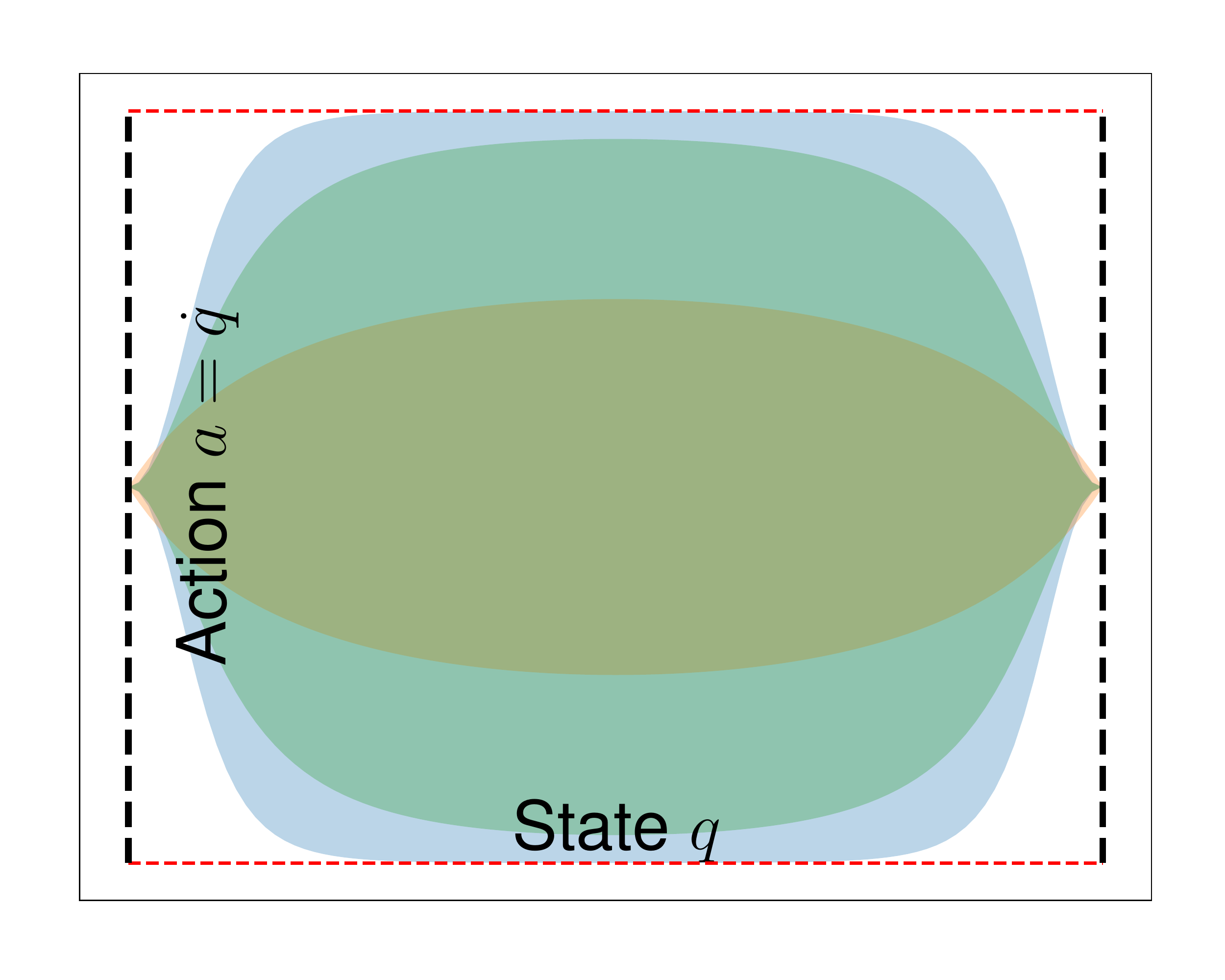}
    \includegraphics[width=0.7\linewidth, trim=18cm 1cm 2cm 2cm, clip]{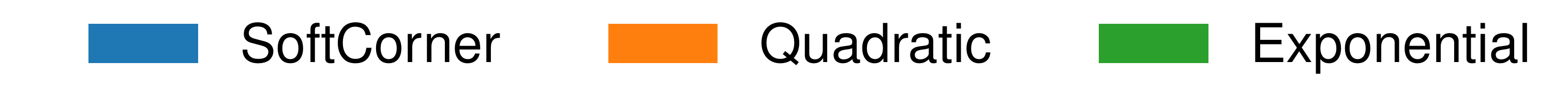}
    \caption{Comparison of different types of slack variables. \textbf{Left:} The constraint manifold defined by different type of slack variables for $q<0$. The quadratic slack variable suffers ambiguity issues at point $(q_0, \mu_1)$ and $(q_0, \mu_2)$. The tangent space bases (orange arrow) leads to opposite directions along $q$-axis. \textbf{Right:} The morphing of the action space with different slacks. The red dashed lines define the original action limits $-1<a<1$ and black dashed lines are constraints $-1<q<0$. The shaded area defines the projected action space for each state $q$ following \cref{eq:bases_projection}}
    \label{fig:slack_types}
\end{figure}

\paragraph{Projected Tangent Space Bases}
The tangent space bases are usually determined by computing the null space of the Jacobian matrix using QR~\cite{kim1986qr} or SVD~\cite{singh1985singular} decomposition. One desired property is to have continuously varying tangent space bases. However, there is no continuous function that generates the null space~\cite{byrd1986continuity}. QR/SVD-based methods use the Householder transformation, which applies a sign  function $sgn(\cdot)$ during the Tridiagonalization. The null space bases can flip the direction when a diagonal element changes sign. However, \cite{byrd1986continuity} proved the continuity of the projection-based method to generate tangent space bases under certain conditions. Following their idea, the projected null space of the Jacobian matrix is determined by 
\begin{equation}
    \scalemath{0.85}{\mN = (\mI - \mJ^{\intercal}(\mJ \mJ^{\intercal})^{-1}\mJ)\mZ} \label{eq:bases_projection}
\end{equation}
where $\mJ$ is the Jacobian of the constraint manifold, $\mZ = [\mI_{n} \; \vzero_k]^{\intercal} \in \RR^{(n+k)\times n}$ are the augmented bases that combine the normalized action-space bases $\mI_{n}\in \RR^{n \times n}$ with $\vzero_n\in \RR^{n \times k}$. \cref{eq:bases_projection} projects the action bases onto the tangent space of the constraint manifold. Notice that the first $n$ dimensions of the augmented state correspond to the original action space, and that the project bases are not orthogonal. 

\paragraph{The SoftCorner Slack Variable}
\label{subsec:softcorner_slack}
\begin{figure*}[!ht]
    \centering
    \includegraphics[width=0.24\textwidth, height=3.5cm, trim=4.5cm 1cm 1cm 1cm, clip]{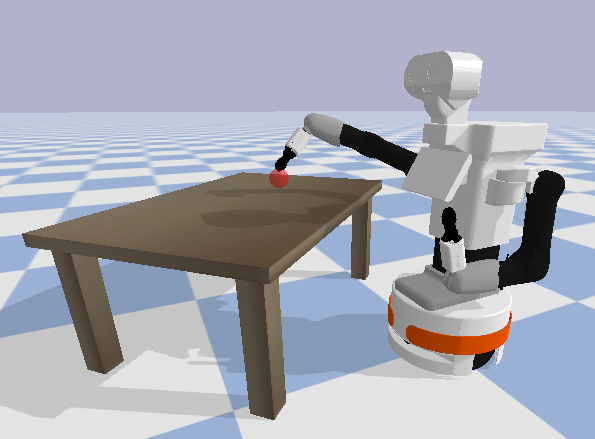}
    \includegraphics[width=0.24\linewidth, height=3.5cm, trim=0.5cm 2cm 1cm 1cm, clip]{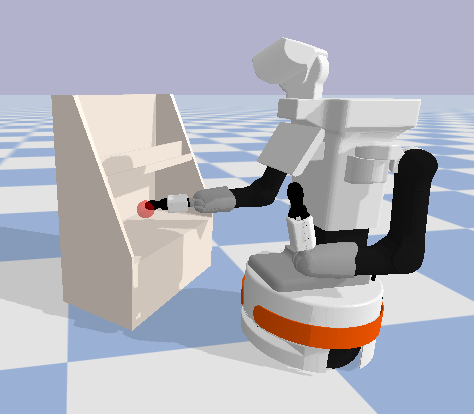}
    \includegraphics[width=0.24\linewidth, height=3.5cm, trim=1cm 2cm 0cm 1cm, clip]{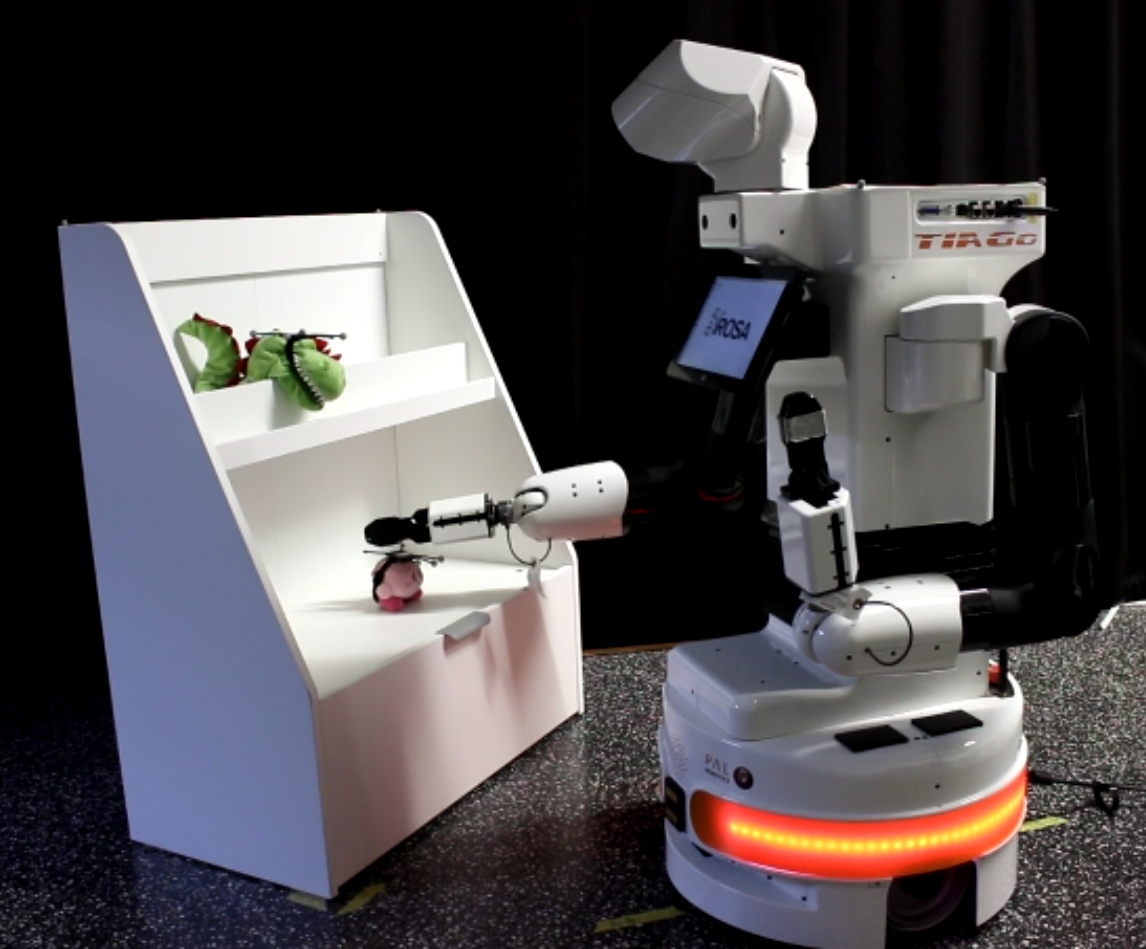}
    \includegraphics[width=0.24\textwidth, height=3.5cm, trim=4cm 3cm 4cm 7cm, clip]{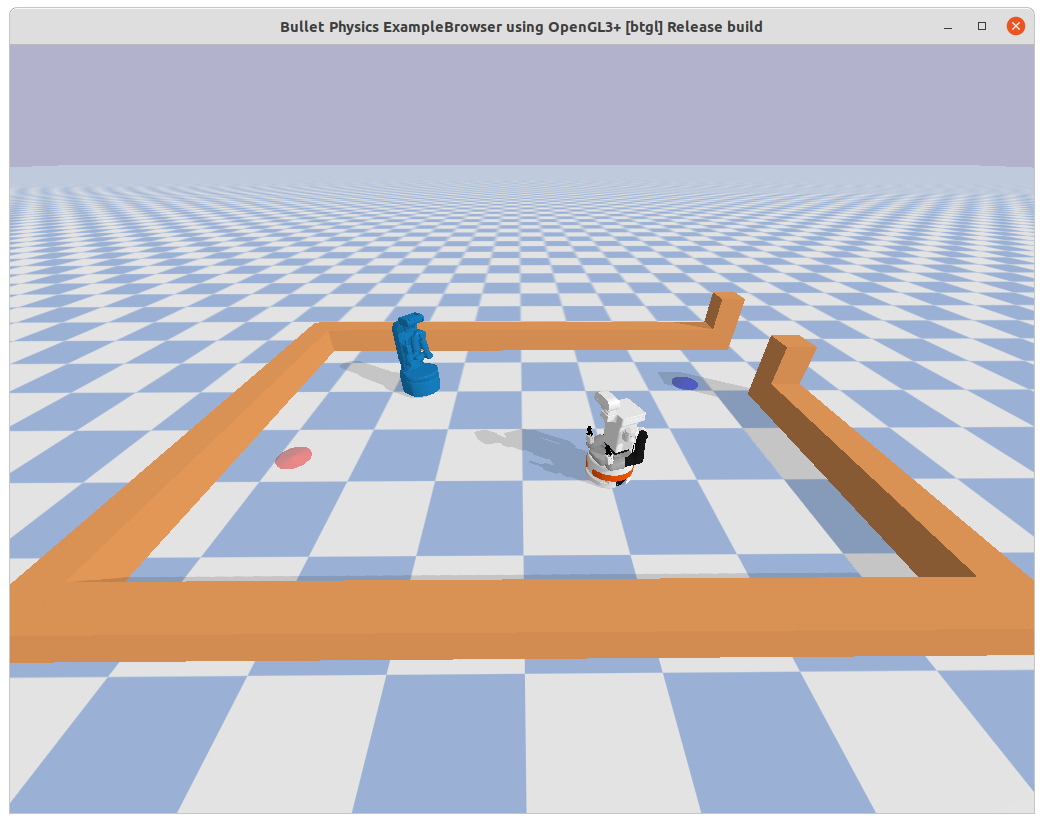}
    \caption{\gls{rl} environments, from left to right: \texttt{TableEnv}, \texttt{ShelfEnvSim}, \texttt{ShelfEnvReal}, \texttt{NavEnv}}
    \vspace{-0.6cm}
    \label{fig:environments}
\end{figure*}
In the original manifold construction, the mapping between the constraint $\vc(\vq)$ and $\vmu$ in \cref{eq:constraint_manifold} is not unique, causing an ambiguity in the tangent space bases. As an example, \cref{fig:slack_types} (right) illustrates the constraint manifold $q + \mu^2 = 0$ for the inequality constraint $q < 0$, where the arrow shows the tangent space basis. The basis of the tangent space at configuration $q_0$ leads to an opposite direction along the $q$-axis because the slack variable $\mu$ is different, as shown by the points $(q_0, \mu_1)$ and $(q_0, \mu_2)$. However, as the slack variable is not observed by the agent, this inconsistency of the basis will cause failures during training. 
Instead, We can formulate a bijective mapping between the original constraint and the slack variable, such as the exponential form $\exp{(\beta \vmu)}$.
Here, $\beta$ is a positive scalar controlling how strong the original action space shrinks as the constraint function approaches the limit: lowering the $\beta$ parameter makes the action space more sensitive to the value of the constraint, as the green curve shown in \cref{fig:slack_types}~(left).
However, this formulation morphs the action space even in states that are far away from constraint violations, as shown in \cref{fig:slack_types}~(right). To avoid this issue, we introduce the \textit{SoftCorner} slack variable parametrization
\begin{equation}
    \scalemath{0.85}{\vc(\vq,\vx) + \dfrac{1}{\beta}\log\left(1-\exp(\beta\vmu)\right) = \vzero},
\end{equation}
where $\beta$ here has the same interpretation as in the exponential parametrization. In addition, the SoftCorner slack variable maintains the original action space when the states are far from violating the constraint. This slack definition is not sensitive to the metric spaces in which the constraints are defined, such as the combination of the joint space and Cartesian space constraints. 

\subsection{Collision Avoidance with Learned SDFs}
Collision avoidance constraints are formulated as maintaining a sufficient distance margin between two objects. Typical approach approximate obstacle with primitive shapes, such as spheres. However, this approach is not feasible for complex or dynamic shapes, such as a shelf or a human, as the computation load grows quadratically with the number of primitives. 
The \gls{sdf} is a prominent representation for expressing distance w.r.t. a given surface by defining a function that precomputes the distance of an arbitrary query point in the Cartesian space.  
As \gls{sdf}s provide a smooth differentiable function, we will discuss how to employ them together with \gls{atacom}.

We rely on~\gls{redsdf}~\cite{liu2022redsdf}, which approximates the distance fields of objects with complex shapes or even articulations, such as humans or robots, and provides distances in a regularized form. \gls{redsdf} uses a neural network to approximate the distance of a query point $\vp$ w.r.t the center of the articulated object, specified by the joint configuration $\vq_{o}$. Unlike DeepSDF~\cite{park2019deepsdf} that focuses on reconstructing the object's surface as close as possible to the zero level-set, the training in \gls{redsdf} integrates an intuitive distance inductive bias, allowing the learning of distance fields at any scale.

The \gls{redsdf} approximates the closest distance of a query point $\vp\in \RR^3$ in Cartesian space w.r.t the object. To integrate the collision avoidance constraints in \gls{atacom}, we define multiple \gls{poi} located at relevant positions of the robot. We can compute the Cartesian position of the \gls{poi}s $\vp_i$ given a robot configuration $\vq$ using forward kinematics $\vp_i = \mathrm{FK}_{i}(\vq), \; i \in [1, 2, ..., N]$. We formulate the collision avoidance constraints as
\begin{equation}
   \scalemath{0.9}{ c_i(\vq, \vq_o): \delta_i - d(\vp_i(\vq), \vq_o) \leq 0, \quad i \in (0, 1, ..., N),} \label{eq:dist_constraint}
\end{equation}
where $\delta_i$ are thresholds assigned for each \gls{poi} and $d(\vp_i(\vq), \vq_o)$ is the distance computed by the \gls{redsdf} model. A key advantage of using \gls{redsdf} is that we can compute the gradient w.r.t the robot configuration $\vq$ as $\nabla_{\vq}c_i = -\nabla_{\vp_i}d \cdot \nabla_{\vq}\vp_i$, with $\nabla_{\vp_i}d$ the gradient of the distance field and $\nabla_{\vq}\vp_i$  the Jacobian of the \gls{poi} w.r.t the robot configuration, which can be computed based on the adjoint matrix~\cite{lynch2017modern}. Note that we don't include the velocity for the distance constraint in \eqref{eq:dist_constraint} as the velocity of the obstacles is considered in \eqref{eq:atacom_affine}.

\section{Experimental Evaluation}
To evaluate the performance of generalized \gls{atacom}, we designed a series of tasks with different complexity. First, we designed static goal-reaching tasks in which the robot has to avoid collisions with objects of different geometric complexity, e.g., a table and a shelf, as shown in \cref{fig:environments}. We then designed two dynamic tasks for navigation and for \gls{hri}. In the robot navigation scenario, we assume a differential-drive mobile robot and test the performance of \gls{atacom} in respecting the control constraints while avoiding collisions with a blindly moving mobile robot. In the \gls{hri} task, we consider a shared-workspace scenario in which a human moves dynamically while the robot attempts to deliver a cup appropriately to the desired goal location without colliding with the human or ``spilling'' the content of the cup. In the following, we provide our empirical evaluation against baselines. All algorithms and environments are implemented within the MushroomRL framework~\cite{deramo2020mushroomrl}; technical and implementation details can be found on the website \url{https://irosalab.com/saferobotrl/}.

\subsection{Manipulation Tasks}\label{subsec:manip}
The first two experiments are goal-reaching tasks in the proximity of the obstacles, defined as a table \texttt{TableEnv} or a shelf \texttt{ShelfEnvSim} (\cref{fig:environments}). To ensure safety, we define 9 query-\gls{poi} distributed along the links of the moving arm for computing the distance constraints with \gls{redsdf} w.r.t. the tables or shelf. We also define distance constraints for avoiding the ground, self-collisions, and joint limits. Both the \texttt{TableEnv} and the \texttt{ShelfEnvSim} contain 34 constraints. The reward is defined as $r(\vs_t, \va_t) = - \rho_d  d_\text{goal} - \rho_{\angle}  d_{\angle} - \rho_a \Vert \va_t \Vert + \mathbb{I}_{\vs}$, with the distance to the goal position $d_\text{goal}$ and orientation $d_{\angle}$, the action penalty $\Vert \va_t \Vert$, the scaling factor $\rho_d, \rho_{\angle}, \rho_a$, and a task indicator $\mathbb{I}_{\vs}$. The task indicator encourages the robot to stay on top of the table or inside the shelf. For the manipulation task, we apply velocity control $\vu = \dot{\vq}$ on the joints with control frequency $30$ Hz. Each episode contains 500 steps. Each episode terminates if a collision is detected, with a penalty of $r_{\mathrm{term}}=-1000$.

\begin{figure}[!tb]
    \centering
    \begin{subfigure}[b]{\linewidth}
         \centering
         \caption{\texttt{TableEnv}}
         \includegraphics[width=\linewidth, trim=0cm 0.7cm 0cm 0.7cm, clip]{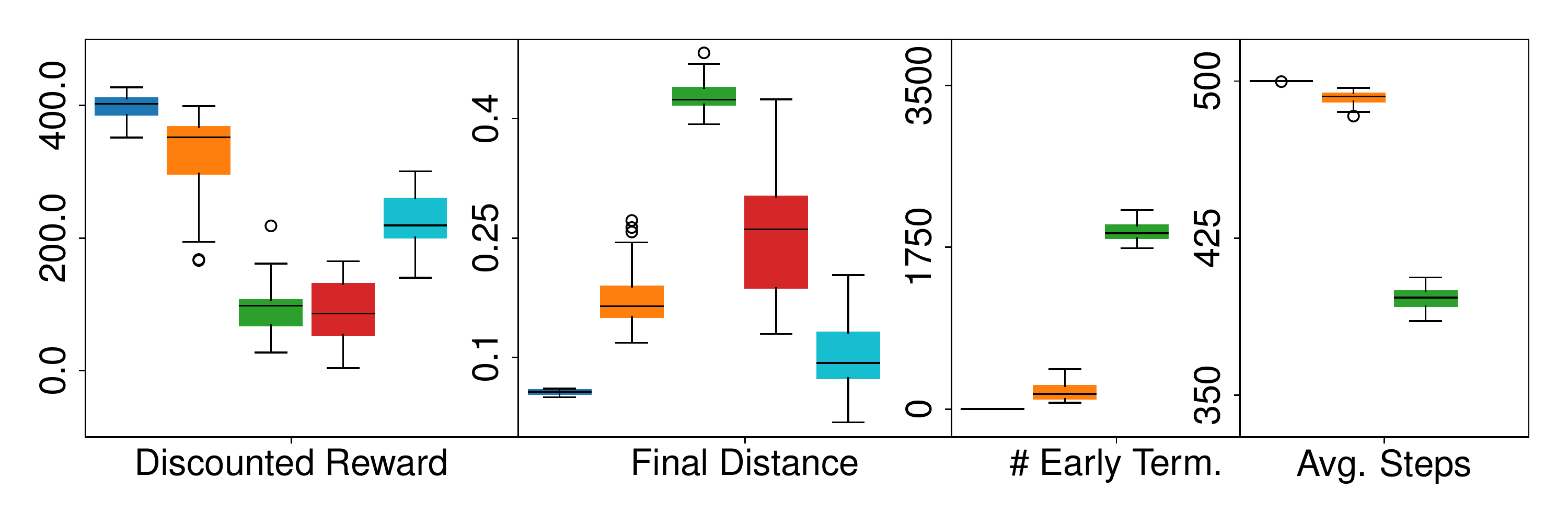} 
         \label{fig:Table_result}
         \vspace{-0.4cm}
    \end{subfigure}
    \begin{subfigure}[b]{\linewidth}
        \centering
        \caption{\texttt{ShelfEnvSim}}
        \includegraphics[width=\linewidth, trim=0cm 0.7cm 0cm 0.7cm, clip]{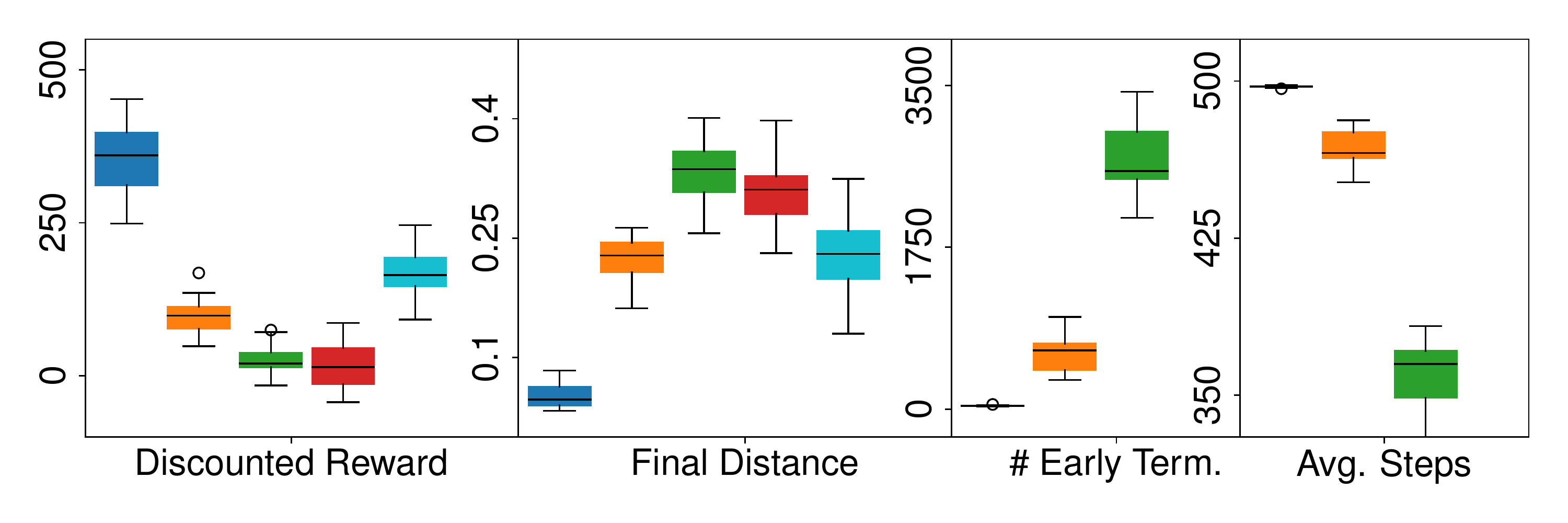}
        \label{fig:Shelf_result}
        \vspace{-0.5cm}
    \end{subfigure}
    \begin{subfigure}[b]{\linewidth}
        \centering
        \caption{\texttt{NavEnv}}
        \includegraphics[width=\linewidth, trim=0cm 0.7cm 0cm 0.7cm, clip]{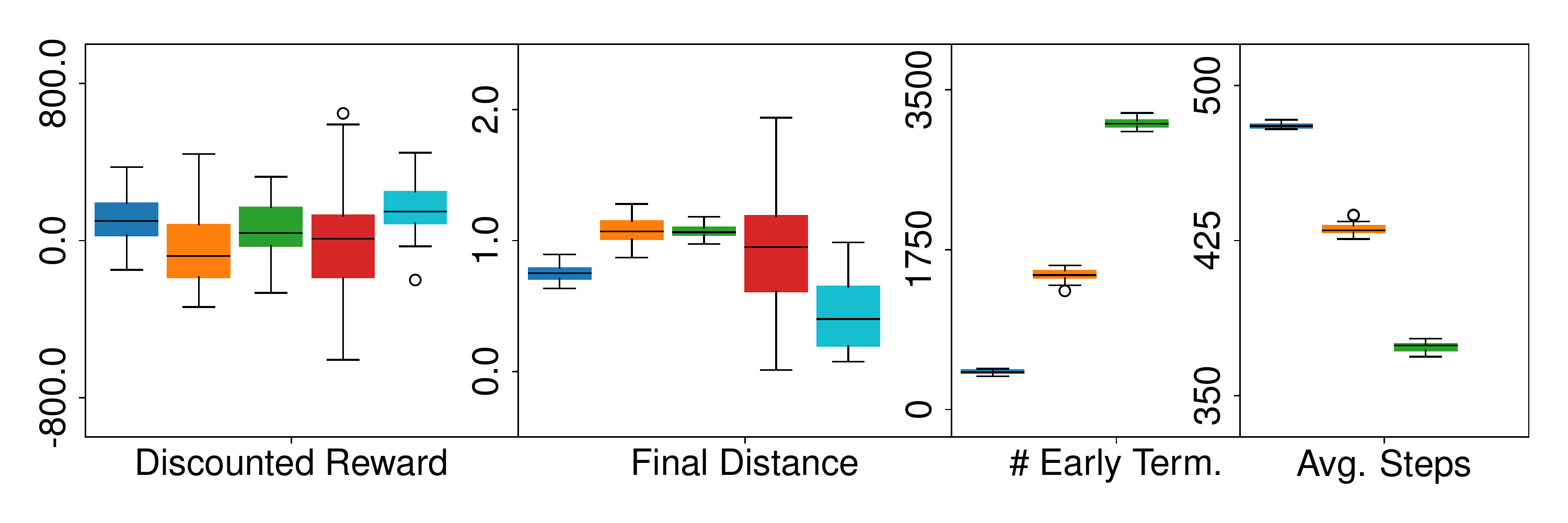}
        \label{fig:Navigation_result}
        \vspace{-0.3cm}
    \end{subfigure}
    \vspace{-0.2cm}
    \includegraphics[width=\linewidth, trim=2cm 2cm 1cm 2cm, clip]{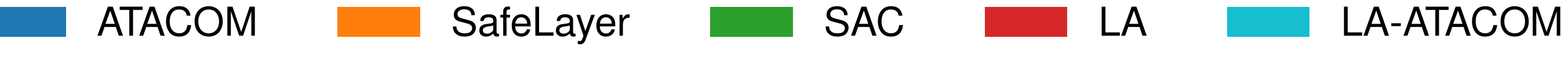}
    \caption{Experimental comparison of \gls{atacom} and baseline methods in different tasks. We report discounted reward, the final distance reached across evaluations, the number of early terminations due to damage events (collision or violation of joint limits), and the average steps per episode (higher means agent living longer through a training episode).}
    \label{fig:box_table_shelf_nav}
    \vspace{-0.8cm}
\end{figure}

We compare our approach with three baselines, i.e., vanilla SAC~\cite{haarnoja2018soft}, SafeLayer~\cite{dalal2018safe}, and a hard-coded Linear Attractor (LA). LA-ATACOM applies ATACOM on top of the action computed by LA. Learning the constraint as proposed in \cite{dalal2018safe} is very challenging; thus, we use the constraint functions defined by our approach for a fair comparison. We conducted a hyperparameter search for each task on the learning rates with 5 random seeds, and then, ran 25 seeds with the best hyperparameters. We report the final discounted reward and the distance to the target before episode termination. We also compare the total number of episodes with early terminations due to damage events (collisions or violation of joint limits) and average episode steps throughout 2 million training steps, as shown in \cref{fig:Table_result} and \cref{fig:Shelf_result}. Training is not required in LA and LA-ATACOM, and only the final performance is reported.
From the boxplots it is clear that learning with \gls{atacom} high-dimensional manipulation tasks achieves better performance, as LA-ATACOM gets stuck in local minima due to joint limits and complex obstacle shapes.


\subsection{Navigation: Differential-Drive Robot}
In the \texttt{NavEnv}, we designed a navigation task for the differential-drive TIAGo++ robot (white) that moves in a room while avoiding the Fetch robot (Blue), as shown in \cref{fig:environments}. The Fetch robot moves to its randomly assigned target goal, using a hand-crafted policy that ignores TIAGo, and serves as a dynamic obstacle. We aim to train TIAGo to reach its randomly generated goal while avoiding collisions with the Fetch and the walls. The control actions are the linear and angular velocities of the robot base. We follow a reward structure similar to the one in \cref{subsec:manip}. However, the orientation reward is defined as $-\mathrm{sigmoid}(30(|d|-0.2))\cdot \frac{|\angle_{\mathrm{goal}} - \theta |}{\pi}$, with the heading angle of the robot $\theta$ and the yaw angle to the goal $\angle_{\mathrm{goal}}$. As shown in \cref{fig:Navigation_result}, the \gls{atacom}-TIAGo can reach the goal with lower error while experiencing significantly fewer collisions. Notably, the performance of SafeLayer in \texttt{NavEnv} is significantly lower compared to the static \texttt{TableEnv} and \texttt{ShelfEnvSim}. Indeed, the SafeLayer approach only corrects the actions when the constraints are violated. To ensure safety in a dynamic task, SafeLayer requires a larger safety threshold, limiting task performance. Instead, \gls{atacom} reduces the feasible action space when approaching the constraint's boundary. In this task, LA-ATACOM outperforms the learned policy as the moving obstacle prevents local minima and the hard-coded policy is a strong and efficient bias to reach a fixed target.

\subsection{Human Robot Interaction}
In the last task, we test our approach in a \texttt{HRI-Sim} environment (\cref{fig:main_figure}) with reward function similar to \cref{subsec:manip}. TIAGo should deliver a cup of water vertically to a target point while the human operates in a shared workspace. Like in \texttt{TableEnv}, we define constraints for avoiding collisions w.r.t the table, the robot, the ground, and the human using the pre-trained \gls{redsdf}.  To simulate human motion, we record a human using a motion capture system ($\sim$18.000 data points). During the recording, the human moves near the table arbitrarily without considering the robot. We convert the human motion to the SMPL representation~\cite{loper2015smpl}. To train a robust policy, we randomly initialize the human motion at the beginning of each episode.  Note that given the inferior performance of the baselines in the previous tasks, we do not compare against them in this far more challenging task to save computational and energy resources. We evaluate the final policy with 10000 steps, the converged \gls{atacom} policy achieves a final distance to target $0.14 \pm 0.02$m and the number of episodes in which a spill happens is $4 \pm 3.16$. Throughout the training process (2 million steps), the total number of collisions were $9.2 \pm 8.11$. The small number of constraint violation during training confirm the effectiveness of the safe exploration strategy of \gls{atacom}, that ends up learning a safe interaction policy.

\subsection{Real Robot Validation}
Given the encouraging results of \gls{atacom} in our simulated results, we transfer our safe policies to the real world\protect\footnote{Videos of the real-world experiments can be found in:\newline
\url{https://irosalab.com/saferobotrl/}}. We create the \texttt{ShelfEnvReal}, in which we place two objects on the shelves of a bookcase (\cref{fig:environments}).  We use motion capture to perceive the objects' poses. The robot starts from a random configuration and reaches the two targets sequentially. The maximum reaching time for each target is 16.67s. Once the distance between the robot grasping frame and the object is smaller than 10cm for 2s, we count the task as a success. We conduct 25 trials with different combinations of object placements and initial arm configuration. As seen in our results in~\cref{tab:real_shelf}, our policy is 100\% safe.

We also conduct real \gls{hri} experiments (\cref{fig:main_figure}). We rely on a motion capture system to get an accurate human pose estimate for querying the human-\gls{redsdf}. In our experimental scenario, the human is instructed to ignore the robot (i.e., the subject looks at the phone) while reaching for an object. The robot, in the meantime, delivers a cup of water while avoiding all possible collisions smoothly, exactly as in the simulated task. We conducted 25 trials in which the human behaved differently. The experiment resulted in a \textbf{minimum distance} of (0.185 $\pm$ 0.021) m a \textbf{success rate} of 96$\%$ and 0 \textbf{collisions}. Note that a task is considered successful when the episode ends with no spills (here, we used granulated sugar instead of water). Our \gls{atacom}-agent trades off a small percentage of task failure to ensure 100\% safety.

\begin{table}[bt!]
    \centering
    \caption{\texttt{ShelfEnvReal} Experiments}
    \begin{tabular}{c|c|c|c|c}
        Target & Final Error (m) & Suc. Rate & $\#$ Collisions & Time (s)\\
        \hline
        1 & 0.038 $\pm$ 0.010 & 90$\%$ & 0 & 6.58 $\pm$ 1.01\\
        2 & 0.047 $\pm$ 0.024 & 90$\%$ & 0 & 6.47 $\pm$ 1.73\\
    \end{tabular}
    \label{tab:real_shelf}
    \vspace{-0.5cm}
\end{table}

\noindent\textbf{Limitations} While, in principle, it would be possible to use \gls{atacom} to train real robots from scratch, it is still problematic for various reasons. First, the high computational complexity of \gls{rl} algorithms for solving challenging robotic tasks would require an impractical amount of time to operate the real robot. Moreover, the complexity of resetting the state of a real robot at the end of each episode renders learning from scratch impractical. Additionally, the Gaussian exploration model may be problematic for robotic actuators. Our approach also has some other practical limitations. To ensure safety, we require a high-performance tracking controller and perfect perception e.g., from motion capture. Furthermore, the computational requirements can become an issue if we use multiple \gls{redsdf} networks to compute the constraints. 
Finally, in this work, we use a simplified model for \gls{hri} i.e., a simulated human unaware of the robot's existence. In reality, the human can behave very differently and would probably try to avoid the robot while interacting with it, at least to some extent.


\section{Conclusions}
This paper introduces a novel and more general formulation of Safe Exploration in \gls{rl} by \glsfirst{atacom}, i.e., by transforming the agent's action space by mapping it on the nullspace of the constraints. We extend the original approach to non-linear control affine systems, hence, treating various robotic platforms and a wide variety of tasks, such as manipulation and navigation. Furthermore, we show how to integrate learned constraints with \gls{redsdf}: this allows imposing arbitrary shapes and articulated structures as safety constraints.
This flexibility in the constraint definition makes learning safe interactions with humans possible, even when deploying the learned policies in the real world, without sacrificing performance. The proposed method paves the way for possible breakthroughs in the deployment of \gls{rl} methods in dynamic real-world tasks. 



\clearpage
\addtolength{\textheight}{-14.5cm} 
\bibliographystyle{IEEEtran}
\bibliography{bibliography}

\end{document}